%% file: acl_latex.tex
\pdfoutput=1

\documentclass[11pt]{article}

\usepackage[final]{acl}

\usepackage{times}
\usepackage{latexsym}

\usepackage[T1]{fontenc}

\usepackage[utf8]{inputenc}

\usepackage{microtype}

\usepackage{inconsolata}

\usepackage{graphicx}
\usepackage{booktabs}
\usepackage{multirow} 
\usepackage{amsmath}    
\usepackage{amsfonts}   
\usepackage{enumitem}   
\usepackage{graphicx}   
\usepackage{float}     
\usepackage{multirow}   
\usepackage{booktabs}   
\usepackage{amssymb}    
\usepackage{tabularx} 
\usepackage{bm} 
\usepackage{tikz}
\usepackage{subcaption}
\usepackage{dashrule}
\usepackage{xcolor}  
\usepackage{pifont}  
\usepackage{xcolor} 
\usepackage{colortbl}  
\usepackage{diagbox} 
\usepackage{graphicx} 
\usepackage{rotating}  

\usepackage{amsmath}   
\usepackage{amsfonts}  
\usepackage{amssymb}   
\usepackage{arydshln} 
\usepackage{tabularx}
\usepackage{hyperref} 
\usepackage{tikz}  
\title{BeMERC: Behavior-Aware MLLM-based Framework for Multimodal Emotion Recognition in Conversation}

\author{
 \textbf{Yumeng Fu},
 \textbf{Junjie Wu},
 \textbf{Zhongjie Wang},
\\
 \textbf{Meishan Zhang},
 \textbf{Yulin Wu},
 \textbf{Bingquan Liu}}

\begin{document}
\maketitle
\begin{abstract}
Multimodal emotion recognition in conversation (MERC), the task of identifying the emotion label for each utterance in a conversation, is vital for developing empathetic machines. Current MLLM-based MERC studies focus mainly on capturing the speaker's textual or vocal characteristics, but ignore the significance of video-derived behavior information. Different from text and audio inputs, learning videos with rich facial expression, body language and posture, provides emotion trigger signals to the models for more accurate emotion predictions. In this paper, we propose a novel behavior-aware MLLM-based framework (BeMERC) to incorporate speaker's behaviors, including subtle facial micro-expression, body language and posture, into a vanilla MLLM-based MERC model, thereby facilitating the modeling of emotional dynamics during a conversation. Furthermore, BeMERC adopts a two-stage instruction tuning strategy to extend the model to the conversations scenario for end-to-end training of a MERC predictor. Experiments demonstrate that BeMERC achieves superior performance than the state-of-the-art methods on two benchmark datasets, and also provides a detailed discussion on the significance of video-derived behavior information in MERC.

\end{abstract}

\section{Introduction}




 ``\textit{One's impression of others is conveyed 7\% through words, 38\% through tone, and 55\% through behavior}'' \citep{mehrabian1971silent}. Multimodal emotion recognition in conversation (MERC) is a fundamental task in the community of natural language processing (NLP), aims to identify emotions for each utterance with multiple modalities during a conversation. Differing from pure texts or audios, videos with linguistic transcriptions contains not only verbal (spoken words) modality, but also non-verbal (acoustic and visual) modalities. It enables dialogue systems to attain abundant information for accurate and nuanced emotion predictions \citep{Ngiam}. There are a series of research works in the realm of MERC, such as modal alignment \citep{li-etal-2023-joyful, shi2023multiemo}, commonsense knowledge enhancement \citep{yang2023bipartite, tu2024multiple} and context modelling \citep{zhang-li-2023-cross, xie-etal-2025-dual}. However, they solely capture the emotional states at utterance level while ignoring the subtle emotional fluctuations in interactive communication.

\begin{figure}[t]
  \includegraphics[width=\linewidth]{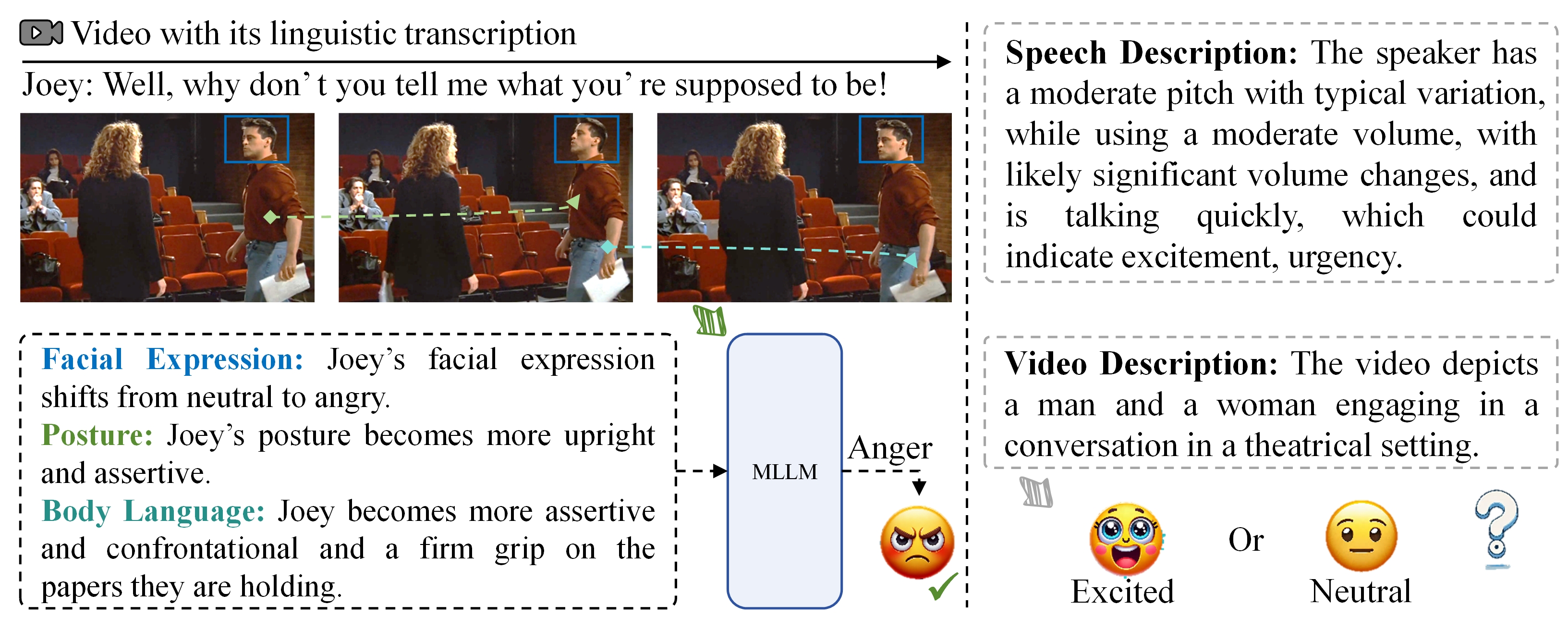}
    \caption{Comparison between video-derived behavior information and other descriptions in the task of MERC. BeMERC integrates facial expression, posture and body language into the MLLM to provide the accurate emotion for the given utterance in a conversation.}
  \label{fig1}
\end{figure}

The key challenge in the realm of MERC stems from emotional dynamics \citep{busso2008iemocap}, referring to emotion trigger signals derived from video data. Advances in MERC have adopted multimodal large language models (MLLMs) to achieve remarkable improvements in performance and generalizability \citep{zhang2023dialoguellm, wu2024beyond}. Despite the non-trivial progress of MLLMs in handling multimodal data, they still remain a significant limitation: \textit{lacking video-derived emotional dynamics, rather than simply introducing speech characteristics or video descriptions}. As illustrated in the right part of Figure~\ref{fig1}, both speech characteristics and video descriptions enable MLLMs-based MERC models to provide inaccurate emotion labels for the target utterance in a conversation.

Based on these preliminary observations and dynamic interaction \citep{9706271} in conversations, we focus on the emotion trigger perspective posed by video contents, which precisely reflects the target emotion of the given utterance. As reported in \citep{mehrabian1971silent}, video-derived behaviors of speaker interaction provides subtle facial micro-expression, body language and posture associated with the target emotion label. Such behavior-aware information can enable MLLM-based MERC models to accurately perform emotion prediction, as depicted in the left part of Figure~\ref{fig1}.

In this paper, to explore the significance of video-derived behavior information in the MERC realm, we present an innovative MLLM-based MERC framework, dubbed BeMERC, designed to capture emotional dynamics for emotion recognition in conversations, as illustrated in Figure~\ref{fig2}. Specifically, on the one hand, to attain video-derived behaviors for emotional dynamics in conversations, we utilize Qwen2-VL \citep{Qwen2-VL} to generate natural language descriptions of subtle facial micro-expression, body language and posture stemming from videos. It facilitates the large language model (LLM) to explicitly associate video-derived behaviors with the emotion of the target utterance. On the other hand, to perform video-derived behaviors incorporation and task-specific tuning, we employ two-stage instruction tuning on the LLM for end-to-end training of a MERC predictor. Empirical results on two widely-used benchmark datasets IEMOCAP and MELD show that the significance of video-derived behavior information in the MERC task, thereby making BeMERC reach a new state-of-the-art (SOTA). We anticipate that this work will provide a sturdy foundation for future MERC research.

In summary, this paper has three-fold contributions as follows:
\begin{itemize}
    \item We introduce BeMERC, a novel behavior-aware MLLM-based framework for emotion recognition in conversations, where emotional dynamics posed by videos can be captured from the emotion trigger perspective. 
    \item We propose a practical way to attain descriptions of video-derived behaviors for triggering the target emotion. For training task-specific BeMERC, we adopt a two-stage instruction tuning strategy that enables the model to learn utterance-level semantic information and subtle emotional fluctuations simultaneously.
    \item Extensive experiments conducted on two benchmark datasets IEMOCAP and MELD demonstrate that the proposed BeMERC significantly surpasses the baseline and SOTA methods in prediction performance.
\end{itemize}

\section{Related Work}
In this section, to underline our contributions, we review existing MERC methods and instruction tuning, respectively.

\subsection{Multimodal Emotion Recognition in Conversations}
Multimodal emotion recognition in conversation (MERC) is one of the most fundamental and important tasks in human daily interactive communication, which requires the extraction and integration of emotional dynamics from conversations. According to feature engineering and architecture design, previous MERC methods \citep{zheng2023facial, chudasama2022m2fnet, ma2023transformer, ai2023gcn, meng2024revisiting, shou2024revisiting} can roughly divided into three types: sequence-based, pre-trained language model-based and graph-based methods. Although these methods attempt to simulate speaker interactions to extract emotional cues for triggering target emotions, they emphasize the semantic relation in the perspectives of utterance or modality, rather than subtle nuances that may be apparent from videos.

The emergence of advanced multimodal large language models (MLLMs) has provided great advancements in numerous multimodal tasks, such as visual question answering \citep{li-etal-2024-groundinggpt} and image caption \citep{jiang2024hallucination}. Due to understanding and generation capabilities of MLLMs, some works have utilized MLLMs to the task of MERC. Typically, SpeechCueLLM \citep{wu2024beyond} translates audios into text descriptions for speech characteristics. Furthermore, DialogueLLM \citep{zhang2023dialoguellm} converts videos into natural language descriptions, which is regarded as supplementary knowledge for emotion analysis. Despite making progress in the realm of MERC, they struggle to capture rich and deep of emotional dynamics from conversations.

In contrast, our work aims to demonstrate that the video-derived behavior information, including subtle facial micro-expression, body language and posture, is significant for the performance of remotion recognition in conversations.

\subsection{Instruction Tuning}
In the context of MLLMs, instruction tuning has emerged as a practice technology to further fine-tuning pre-trained models with instruction-following datasets for diverse downstream tasks \citep{brown2020language, InstructBLIP, qi2024sniffer, cheng2024emotion}. Studies like GroundingGPT \citep{li-etal-2024-groundinggpt}, InstructERC \citep{lei2023instructerc}, and DialogueLLM \citep{zhang2023dialoguellm} have investigated instruction-tuning methods that significantly provide substantial performance improvements in unseen tasks. One challenge of instruction tuning in the task of MERC is how to attain high-quality instructions that refer to the video-derived behaviors for triggering the target emotions in conversations. Therefore, in this work, we extend the capability of general-purpose MLLM for the MERC task via instruction tuning.

\section{Method}
\begin{figure*}[t!]
\centering
\includegraphics[width=1\linewidth]{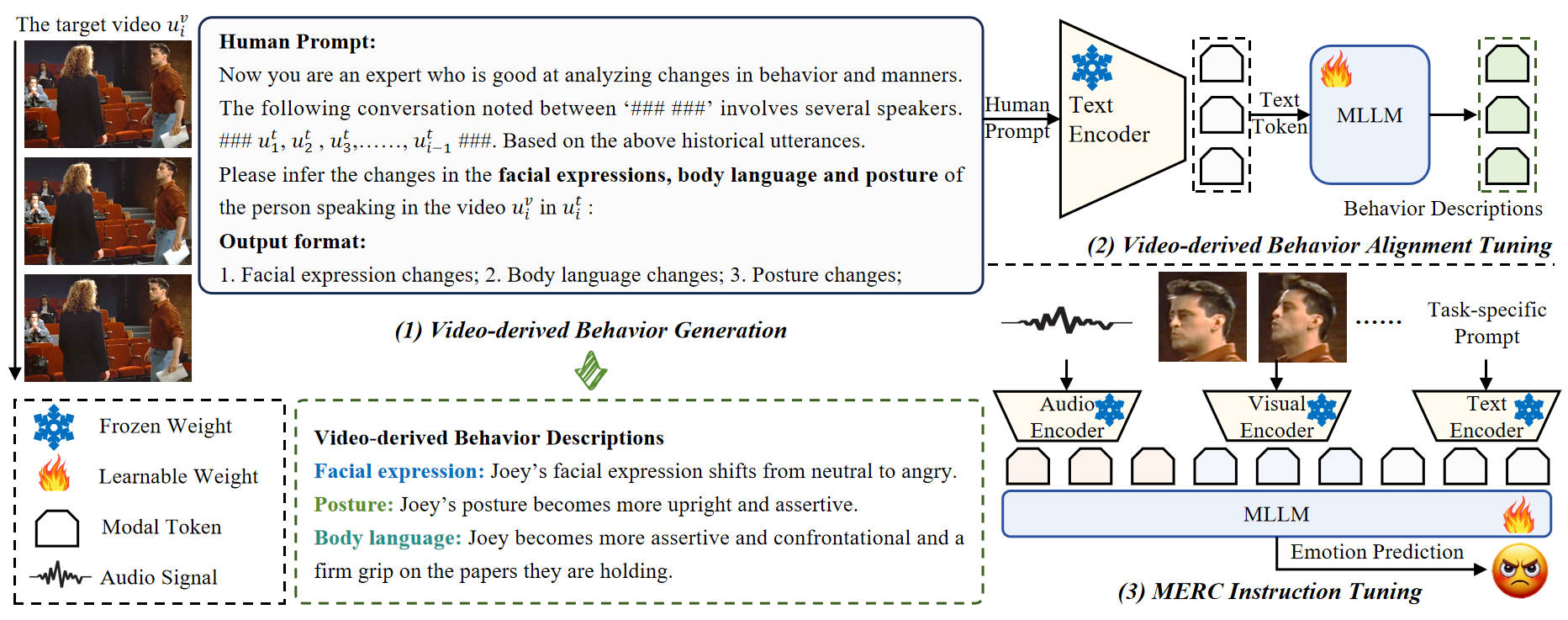}
\caption{The overview of BeMERC. BeMERC includes video-derived behavior generation, video-derived behavior alignment tuning and MERC instruction tuning. In video-derived behavior alignment tuning stage, the generated behaviors are employed to enhance the LLM perceiving emotional dynamics. In the MERC instruction tuning  stage, it improves the ability of the LLM to perform MERC tasks.}
%
\label{fig2}

\end{figure*}

In this section, we propose a framework, namely BeMERC, which enhances the performance of LLMs in ERC tasks by capturing speaker dynamics within utterance through the modeling of three different speaker behaviors.
Figure~\ref{fig2} presents an overview of BeMERC. 
It consists of three steps, including video-derived behavior generation, video-derived behavior alignment tuning and MERC instruction tuning. 

\subsection{Vanilla MERC Model}
A conversation data source can be denoted as $\mathcal{D}=\left\{C_i\right\}_{i=1}^N$, where $C_i$ represents the $i$-th conversation, and $N$ is the size of $\mathcal{D}$. 
Each conversation includes a sequence of utterances $\mathcal{U}=\left\{u_j\right\}_{j=1}^S$, where $S$ is the number of all utterances. 
Each utterance $u_j=(a_j,v_j,t_j,y_j)$, where $a_j$, $v_j$ and $t_j$ respectively represent information from audo, video and text modalities, as well as  $y_j\in\left\{e_1,e_2,...,e_K\right\}$, and $K$ is the number of emotion categories.

Generally, the MERC model $\mathcal{M}$ based on MLLMs is learned from $\mathcal{D}$ to provide a response $r$ over a set of the predefined emotion labels $\mathcal{E}=\left\{e_k\right\}_{k=1}^K$. The whole process can be expressed as follows:
\begin{eqnarray}\label{eq:vcg_o}
  r_{j,i} = \mathcal{M}(u_{<j}, u_j, \mathcal{E}),
\end{eqnarray}
\noindent where, $u_{<j}$ denotes the historical utterances before the target utterance $u_j$ in the $i$-th conversation.


\subsection{Video-derived Behavior Generation}
In our paper, the behavior information of the speaker includes facial expressions, body language, and posture \citep{mehrabian1971silent}. Among them, body language reflects the intention behind an action, facial expressions convey the mental state, and posture reveals the attitude toward an event \citep{sinke2013body}. Since they all capture fine-grained emotional dynamics (Sinke et al., 2013), thus modeling these behaviors helps enhance the performance on the MERC tasks

Considering that large language models possess extensive world knowledge 
and text can be expressed in a clear and simple dynamics form, we design a template to leverage the more powerful model for extracting textual descriptions of behavior changes from videos, including facial expression, body language, and posture.
We utilize knowledge distillation techniques to transfer the perceptual capabilities of a heavyweight general LLM to a lightweight specialized LLM designed for ERC tasks, as shown in Figure~\ref{fig2}(a). 
Specifically, we start by using the text modality prompt $T_{b}$ (Prompt is described in \S\ref{ape_tb}) and multimodal conversation information $\{v_i,a_i,t_i\}$ to construct a complete multimodal prompt $\tilde{T}_{b}$. 
Next, we feed $\tilde{T}_{b}$ into a heavyweight LLM to obtain three different behaviors.
%
%
%
Additionally, followed by~\citep{lei2023instructerc}, we reduce the impact of prompts randomness by structuring our prompt.

\subsection{Video-derived Behavior Alignment Tuning}


The purpose of this step is to use the behavior information obtained from the previous stage as pseudo-labels for training a lightweight behavior-aware LLM. 
%
%
Unlike the previous stage, when constructing multimodal prompt, we only use the text modality, while the tokens corresponding to the remaining modalities are replaced with predefined placeholders. 
This requires the LLMs to infer human behavior with limited available information, thereby further enhancing its perception of behavior. 
As for those placeholders, they will be replaced in the next stage by actual video and audio tokens, thereby equipping our LLM with the ability to process multimodal information.

\subsection{MERC Instruction Tuning}
\label{sub:multimodal_emotion_recognition}
This step is responsible for two purposes. One is enhancing the emotion recognition capabilities of the LLM. Another is to enable the LLM to process multimodal information, thereby fully unlocking its potential to perceive behaviors. 
To achieve these, we first design a new structured template $T_e$ (Prompt is described in \S\ref{ape_tb}), which requires the LLM to predict emotion label for specific utterance. 
However, the template $T_e$ is not used directly, instead, it will be concatenated with the tokens corresponding to the video and audio modalities before being fed into the LLM, which allows the LLM to simultaneously acquire the ability to judge emotions and process multimodal information.
We use DenseNet~\citep{huang2017densely} and OpenSMILE~\citep{eyben2010opensmile} as our video and audio encoders. 
Specifically, for the video modality, we randomly sample 64 frames of images from each video and resize the shape of images to 336*336 before feeding them into DenseNet for encoding. 
For the audio modality, we extract frames every two seconds and send them into OpenSMILE for encoding. 
The encoding results are then passed through corresponding adapters, consisting of fully connected layers, for dimensional unifying, so that they can be concatenated with the text tokens to construct the complete multimodal template. 
We input the complete prompts into the LLMs for prediction.
Afterward, we optimize the LLM by comparing the prediction results with the their actual emotion labels.
The overall training objective involved in this step is shown in E.q~(\ref{eq:1}), where $\hat{y}$ represents the prediction result and $\lambda \|W\|_2^2$ are L2-regularization.

\begin{equation}\label{eq:1}
    \begin{aligned}
\mathcal{L} = - \sum_{i=1}^{N} \sum_{k=1}^{K}  y_{i,k} \log(\hat{y}_{i,k}) + \lambda \|W\|_2^2
    \end{aligned}
\end{equation}

\section{Experiments}

\input{table1.tex}
\input{table2.tex}

In this section, we present two widely-used conversational datasets, evaluation metrics, compared methods, implementation details of BeMERC and subsequently conduct a thorough analysis of the experimental outcomes.

\subsection{Datasets and Evaluation Metrics}
We use the following two standard datasets for all the experiments:
IEMOCAP \citep{busso2008iemocap}, and MELD \citep{poria2018meld}. More details are shown in Appendix~\ref{A.1}. Following previous works \citep{ma2023transformer}, we present the accuracy (Acc.) and F1 score of the proposed method, along with those of other baseline methods, for each emotion category. We also report the overall accuracy (Acc.) and weighted average F1 score (w-F1). 

\subsection{Compared Methods}
We compare BeMERC with the following models: ICON \citep{hazarika2018icon}, DialogueRNN \citep{majumder2019dialoguernn}, MMGCN \citep{hu2021mmgcn}, DialogueTRM \citep{mao2021dialoguetrm}, MM-DFN \citep{hu2022mm}, UniMSE \citep{hu2022unimse}, SDT \citep{ma2023transformer}, FacialMMT\citep{zheng2023facial}, GS-MCC \citep{meng2024revisiting}, M3Net \citep{chen2023multivariate}, MultiEMO \citep{shi2023multiemo}, SpeechCueLLM 
\citep{wu2024beyond}, DGODE \citep{shou2025dynamic}

\input{table4.tex}

\subsection{Implementation Details}
In the experiments, we adopt Qwen2-VL \citep{wang2024qwen2} to generate speaker interaction information. Also, we use Vicuna-v1.5-7B \citep{chiang2023vicuna} as our foundation model. In order to reduce the trainable parameters of the model while  minimizing the loss of model performance, we use LoRA \citep{hu2021lora} to fine-tune large language model. In hyperparameter settings, we set the learning-rate to 2e-4. For the hyper-parameter such as epoch, we tune them on the development dataset. The reported results are an average over five random runs. Our experimental results are statistically significant under paired t-test (all $p<0.05$). We train with FP16 precision on 2 × 40G Nvidia A100 GPUs.

\subsection{Main Results}
To demonstrate the effectiveness of the proposed BeMERC in the task MERC, we present a comparison with methods on IEMOCAP and MELD datasets, as shown in Table~\ref{tab1} and Table~\ref{tab2}. We can observe that our proposed BeMERC achieves the best performance compared to all methods with respect to overall accuracy and weighted F1-score on the two datasets. This performance underscores that BeMERC exhibits superior generalization ability and yields more precise predictions in the context of emotion recognition.

Generally, compared to previous MERC paradigms, methods based on LLMs have delivered more significant results, benefiting from the generalization and 
discriminative capacities of large language models. Notably, our proposed method BeMERC achieves an improvement of 2.28\% on IEMOCAP dataset and 2.18\% on MELD datasets over SpeechCueLLM. Additionally, compared to our proposed LLM-based baseline LLMERC, BeMERC achieves 74.88\% on IEMOCAP dataset, surpassing LLMERC by 3.37\%. Even in the complex conversation scenarios MELD dataset, BeMERC can also achieve 69.78\% and outperforms LLMERC by  0.88\%. This is due to the efficiency of behavior information in the proposed BeMERC.

Moreover, we can observe an overall increase in sentiment across most categories, with the exception of a slight decline in certain label such as "sad." Upon closer examination of the data, it is noted that speakers associated with this label inherently possess a calm and composed demeanor, typically exhibiting minimal facial expression changes. This characteristic may contribute to minor inaccuracies in recognition outcomes.

\subsection{Abalation Study}

\noindent\textbf{Impact of Behavior Categories.} 
To validate the effectiveness of different types of behavior, we present the results of relevant experiments in Table~\ref{tab4}. It can be observed that the removal of various combinations of behavior types led to varying degrees of performance decline. Among the three types of behavior, the removal of facial expression results in the most noticeable decline. It can be attributed to the emotional cues embedded in the extracted facial expressions, which contribute to improving the accuracy of multimodal dialogue emotion recognition. Additionally, the results show that the removal of all three types of behavioral interaction information resulted in the most significant decline, with a 3.02\% change observed on the IEMOCAP dataset, due to the complementary nature of the behavior interaction information. These experiments demonstrate that BeMERC can achieve precise emotion prediction by incorporating behavior information.


\section{In-depth Analysis}

\subsection{Video-derived Behavior Information
Enriches the Emotion Clues}

\input{leida.tex}
\input{leida2.tex}

In the conversation, we can observe the first utterance in the conversation, such as "Hello". The emotion  of such utterance is inconsistent across different contexts. However, we do not have the preceding context, it is difficult to accurately predict the emotion, often resulting in a default prediction of `Neutral' Therefore, in Figure~\ref{leida}, after incorporating behavior information, we observe a reduction amount in neutral label prediction, indicating the effectiveness of behavior information. However, Figure~\ref{leida} shows that the change in sample distribution after introducing behavior information to the MELD dataset is not very noticeable. This is because the sample distribution of the MELD dataset is imbalanced, with the `Neutral' label accounting for as much as 50\% of the dataset.

\subsection{Different Behavior Information Conveys
Consistent Emotion Clues}
As shown in Figure~\ref{leida2}, we analyze the emotion recognition performance based on any one of the three different behavior information across various emotion labels. 
The results indicate that for all emotion labels, the performance of emotion recognition achieved using different behavior information is consistent. 
This suggests that the emotion clues conveyed by each type of behavior information are aligned, thereby verifying the reliability of our extracted behavioral information.

\subsection{Qualitative Analysis }
\input{julei}
To better illustrate the classification performance of the BeMERC method in the MERC task, we employ principal components analysis to visualize the generated embeddings. As depicted in Figure~\ref{julei}, the BeMERC model exhibits strong performance on the IEMOCAP dataset, with clear separation of samples belonging to different emotional categories. In comparison, while the model without behavior is able to partially distinguish samples across emotional categories, its classification performance is inferior to that of BeMERC. The sample distribution from the model without behavior appears more disorganized, and the boundaries between emotional categories remain ambiguous.

 \subsection{Experiment on Different LLMs}

\input{table6}

To validate the scalability of speaker behavior information across various scenarios and large language models. Specifically, we employ a series of representative LLMs, including GPT-4o-mini \citep{hurst2024gpt}, Gemini \citep{team2023gemini}, Qwen-2.5-72B \citep{yang2024qwen2}, Phi-4-14B \citep{abdin2024phi}, for evaluation in zero-shot scenario. In Table~\ref{tab6},  we can find that various LLMs generate the speaker behavior information that is beneficial to provide the performance improvements of the LLMs-based MERC. However, The lower performance of LLMs in zero-shot than instruction-tuning may be due to a mismatch between their interpretations and the specific labeling 
protocols of datasets \citep{zhao2023chatgpt}. It further reinforces the necessity of our approach in the instruction-tuning scenario.

\section{Case Study}

\begin{figure}[t]
    \centering
  \includegraphics[width=0.9\linewidth]
  {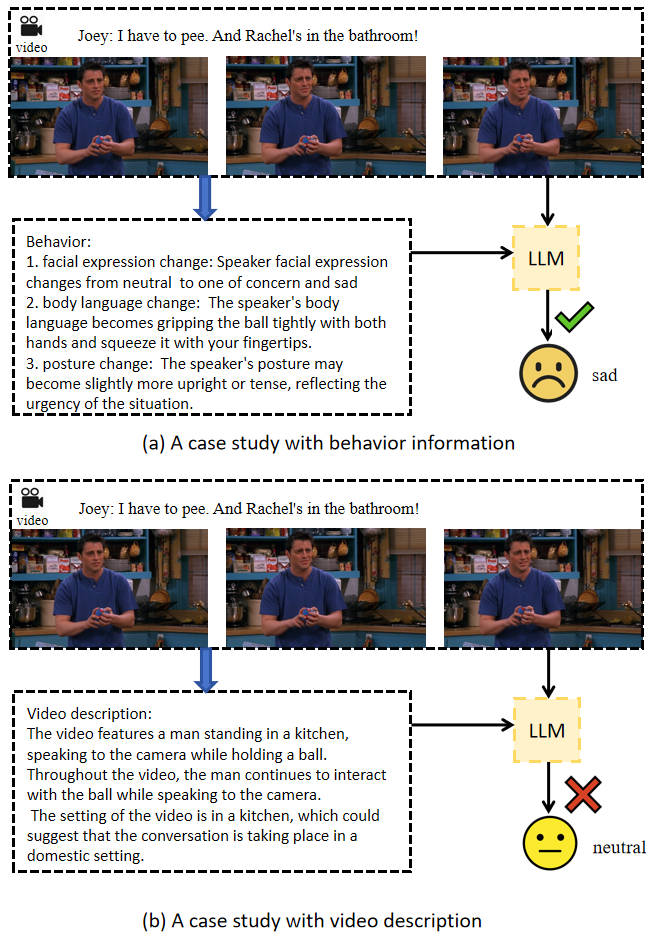}
  \caption {Two cases on meld dataset.}
  \label{case}
\end{figure}

In this section, we present behavior influence to MERC.
Figure~\ref{case} (a) and (b) gives two demonstrations from MELD dataset about the same sentence affected by using behavior information and video description to generate different emotional responses . Conversation (a) predict a accurate label due to the fact that 
behavior is expressed with emotional dynamic clues.   

\section{Conclusion}

In this paper, we propose BeMERC, a novel framework that explores behavior, including facial expression, body language and posture, to promote the progress of multimodal emotion recognition in conversation (MERC). BeMERC is well-designed with three imperative parts, all of which work in harmony to make the model reason emotional dynamics and identify emotional tendencies for each utterance in conversations. Also, BeMERC has verified Mehrabian's Rule.

\bibliography{custom}

\appendix

\section{Example Appendix}
\subsection{Prompts definition}
\label{ape_tb}
\begin{figure}[t]
    \centering
  \includegraphics[width=0.7\linewidth]
  {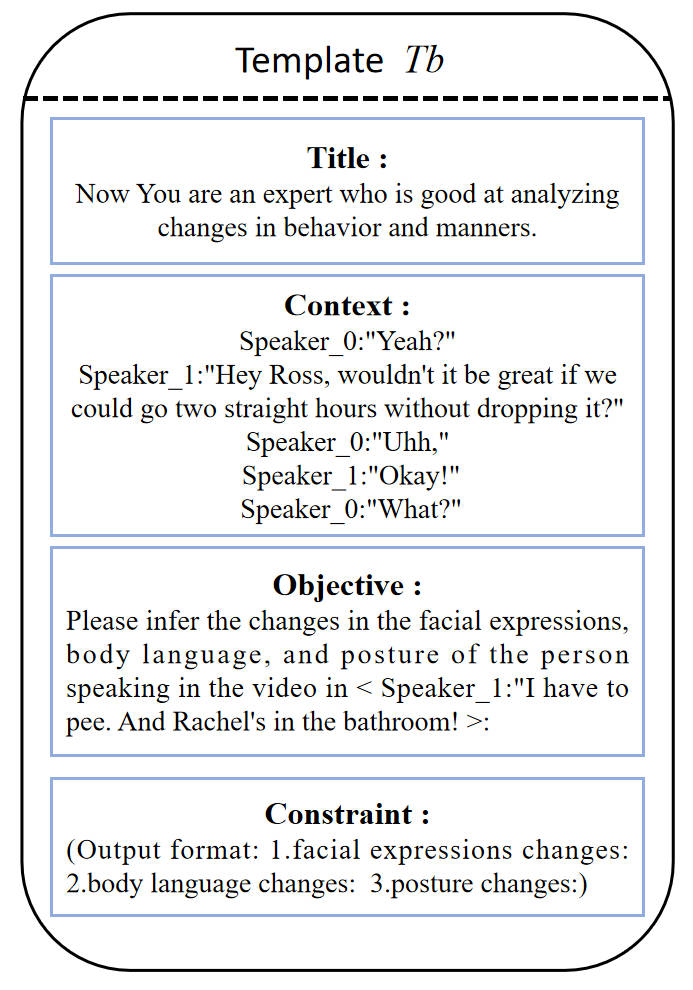}
  \caption {A prompt for video-derived behavior generation and video-derived behavior alignment tuning.}
  \label{template_tb}
\end{figure}

\begin{figure}[t]
    \centering
  \includegraphics[width=0.7\linewidth]
  {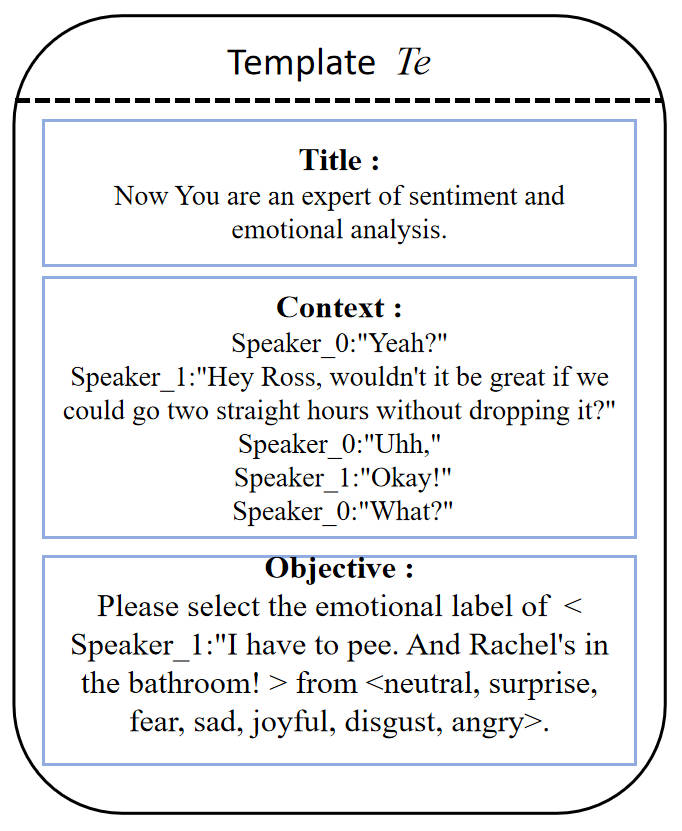}
  \caption {A prompt for MERC Instruction Tuning.}
  \label{template_te}
\end{figure}

From Figure ~\ref{template_tb} and Figure ~\ref{template_te}, we can see the template includes `title', `context', `objective', and `constraint'. The `title' indicates that the role of LLMs expert apt in learning emotional clues in conversations. The `context' give history information of the conversation. The `objective' refers to a concise elucidation of the tasks. The `constraint' is used to restrict the output format.
\label{A.10}

\end{document}

%% file: table1.tex

\begin{table*}[t!]
\renewcommand{\arraystretch}{1}
\setlength{\tabcolsep}{3pt}  
\centering
\resizebox{1.0\linewidth}{!}
{
\begin{tabular}{l|ccccccccccccccccccccc}
\toprule
\multirow{3}{*}{Methods} &  & \multicolumn{20}{c}{IEMOCAP}                                                                                                                                                                                                    \\ \cline{3-22} 
                         &  & \multicolumn{2}{c}{Happy} &  & \multicolumn{2}{c}{Sad} &  & \multicolumn{2}{c}{Neutral} &  & \multicolumn{2}{c}{Angry} &  & \multicolumn{2}{c}{Excited} &  & \multicolumn{2}{c}{Frustrated} &  & \multicolumn{2}{c}{Average} \\
                         &  & Acc.        & F1          &  & Acc.       & F1         &  & Acc.         & F1           &  & Acc.        & F1          &  & Acc.         & F1           &  & Acc.           & F1            &  & Acc.           & w-F1            \\ \midrule
ICON      &  & 25.00 & 31.30  &  & 67.35 & 73.17                 &  & 55.99 & 58.50  &  & 69.41 & 66.29                 &  & 70.90 & 67.09  &  & 71.92 & 65.08                 &  & 62.85 & 62.25          \\
DialogueRNN&  & 25.69 & 33.18  &  & 75.10 & 78.80                 &  & 58.59 & 59.21  &  & 64.71 & 65.28                 &  & 80.27 & 71.86  &  & 61.15 & 58.91                 &  & 63.40 & 62.75          \\ 
MMGCN &  & 32.64 & 39.66  &  & 72.65 & 76.89                 &  & 65.10 & 62.81  &  & 73.53 & 71.43                 &  & 77.93 & 75.40  &  & 65.09 & 63.43                 &  & 66.61 & 66.25          \\
DialogueTRM &  & 61.11 & 57.89  &  &84.90 & 81.25                 &  & 69.27 & 68.56  &  & 76.47 & 75.99                 &  & 76.25 & 76.13  &  & 50.39 & 58.09                 &  & 68.52 & 68.20          \\
MM-DFN &  & 44.44 & 44.44  &  & 77.55 & 80.00                 &  & 71.35 & 66.99  &  & 75.88 & 70.88                 &  & 74.25 & 76.42  &  & 58.27 & 61.67                 &  & 67.84 & 67.85          \\
UniMSE &  & - & -  &  & - & -                 &  & - & -  &  & - & -                 &  & - & -  &  & - & -                 &  & 70.56 & 70.66          \\

SDT &  & 72.71 & 66.19  &  & 79.51 & 81.84                 &  & 76.33 & 74.62  &  & 71.88 & 69.73                 &  & 76.79 & 80.17  &  & 67.14 & 68.68                &  & 73.95 & 74.08          \\

GS-MCC &  & 60.20 & 65.40  &  & 86.20 & 81.20                 &  & 75.70 & 70.90  &  & 71.70 & 70.80                 &  & 83.20 & 81.40  &  & 66.00 & 71.00                
&  & 73.80 & 73.90          \\

DGODE &  & - & 71.80  &  & - & 71.00                 &  & - & 74.90  &  & - & 55.70                 &  & - & 78.60  &  & - & 75.20                
&  & - & 72.80          \\

M3Net &  & - & -  &  & - & -                 &  & - & -  &  & - & -                 &  & - & -  &  & - & -                 &  & 72.46 & 72.49          \\


MultiEMO &  & - & 65.77  &  & - & 85.49                 &  & - & 67.08  &  & - & 69.88                 &  & - & 77.31  &  & - & 70.98                
&  & - & 72.84          \\

SpeechCueLLM &  & - & -  &  & - & -                 &  & - & -  &  & - & -                 &  & - & -  &  & - & -                 &  & - & 72.6          \\ \hdashline

LLMERC
                &  & 58.33        & 57.14        &  & 87.76       & 81.29       &  & 75.00         & 73.94         &  & 80.59        & 69.90       &  & 69.90         & 74.78         &  & 60.37           & 66.38          &  & 71.66           & 71.51          \\ 
BeMERC
                &  & 56.94        & \textbf{59.42}        &  & \textbf{88.16}       & \textbf{84.54}       &  & 72.66         & \textbf{75.20}         &  & 70.00        & 69.39       &  & \textbf{80.27}        & \textbf{79.87}         &  & \textbf{73.75}           & \textbf{72.70}          &  & \textbf{74.98}           & \textbf{74.88}         \\ \bottomrule
\end{tabular}
}
\caption{Performance comparison between our proposed BeMERC and existing MERC methods on IEMOCAP dataset. The experimental results of all the comparative experiments are derived from previously published articles. LLMERC (only use MERC Instruction Tuning stage) which we propose is the baseline for multimodal emotion recognition in conversation based on large language model. The best results are bolded compared with the baseline LLMERC. `-' means that the results are unavailable from the original paper.}
\label{tab1}

\end{table*}

%% file: table2.tex

\begin{table*}[t!]
\renewcommand{\arraystretch}{1}
\setlength{\tabcolsep}{3pt}  
\centering
\resizebox{1.0\linewidth}{!}
{
\begin{tabular}{l|cccccccccccccccccccccccc}
\toprule
\multirow{3}{*}{Methods} &  & \multicolumn{22}{c}{MELD}                                                                                                                                                                                                    \\ \cline{3-25} 
                         &  & \multicolumn{2}{c}{Neutral} &  & \multicolumn{2}{c}{Surprise} &  & \multicolumn{2}{c}{Fear} &  & \multicolumn{2}{c}{Sad} &  & \multicolumn{2}{c}{Joy} &  & \multicolumn{2}{c}{Disgust} &  & \multicolumn{2}{c}{Anger}  &  & \multicolumn{2}{c}{Average} \\
                         &  & Acc.        & F1          &  & Acc.       & F1         &  & Acc.         & F1           &  & Acc.        & F1          &  & Acc.         & F1           &  & Acc.           & F1  
                          &  & Acc.           & F1  
                          &  & Acc.           & w-F1            \\ \midrule
DialogueRNN &  & 82.17 & 76.56  
            &  & 46.62 & 47.64                 &  & 0.00 & 0.00  
            &  & 21.15 & 24.65                 &  & 49.50 & 51.49  
            &  & 0.00 & 0.00                 &  & 48.41 & 46.01      
            &  & 60.27 & 57.95    \\

MMGCN &  & 84.32 & 76.96  
            &  & 47.33 & 49.63                 &  & 2.00 & 3.64  
            &  & 14.90 & 20.39                 &  & 56.97 & 53.76  
            &  & 1.47 & 2.82                 &  & 42.61 & 45.23      
            &  & 61.34 & 58.41    \\

DialogueTRM &  & 83.20 & 79.41  
            &  & 56.94 & 55.27                 &  & 12.00 & 17.39  
            &  & 27.88 & 36.48                 &  & 60.45 & 60.30  
            &  & 16.18 & 20.18                 &  & 51.01 & 49.79      
            &  & 65.10 & 63.80    \\

MM-DFN &  & 79.06 & 75.80  
            &  & 53.02 & 50.42                 &  & 0.00 & 0.00  
            &  & 17.79 & 23.72                 &  & 59.20 & 55.48  
            &  & 0.00 & 0.00                 &  & 50.43 & 48.27      
            &  & 60.96 & 58.72    \\
UniMSE &  & - & -  
            &  & - & - 
            &  & - & -
            &  & - & - 
            &  & - & -
            &  & - & - 
            &  & - & -
            &  & 65.09 & 65.51    \\

SDT &  & 83.22 & 80.19  
            &  & 61.28 & 59.07 
            &  & 13.80 & 17.88
            &  & 34.90 & 43.69 
            &  & 63.24 & 64.29
            &  & 22.65 & 28.78 
            &  & 56.93 & 54.33
            &  & 67.55 & 66.60    \\
FacialMMT &  & - & 80.13  
            &  & - & 59.63 
            &  & - & 19.18
            &  & - & 41.99
            &  & - & 64.88
            &  & - & 18.18 
            &  & - & 56.00
            &  & - & 66.58    \\

DGODE &  & - & 82.60  
            &  & - & 60.90 
            &  & - & 5.10
            &  & - & 45.50 
            &  & - & 63.40
            &  & - & 10.60 
            &  & - & 54.00
            &  & - & 67.20    \\

M3Net &  & - & -  
            &  & - & - 
            &  & - & -
            &  & - & - 
            &  & - & -
            &  & - & - 
            &  & - & -
            &  &  68.28 & 67.05    \\

MultiEMO &  & - & 79.95  
            &  & - & 60.98 
            &  & - & 29.67
            &  & - & 41.51 
            &  & - & 62.82
            &  & - & 36.75 
            &  & - & 54.41
            &  &  - & 66.74    \\
            
SpeechCueLLM &  & - & -  &  & - & -                 &  & - & -  &  & - & -                 &  & - & -  &  & - & -                 &  & - & -       &  & - & 67.60    \\ \hdashline

LLMERC &  & 87.18 & 82.02  
            &  & 64.05 & 61.75 
            &  & 20.00 & 25.97
            &  & 35.57 & 45.82 
            &  & 63.43 & 66.23
            &  & 17.64 & 26.37 
            &  & 60.00 & 58.64
            &  &  70.22 & 68.90    \\

BeMERC &  & \textbf{87.90} & \textbf{82.33}  
            &  & \textbf{63.35} & \textbf{64.49} 
            &  & \textbf{26.00} & \textbf{32.10}
            &  & 33.65 & 44.03 
            &  & \textbf{66.67} & \textbf{66.83}
            &  & \textbf{19.12} & \textbf{30.23}
            &  & \textbf{61.45} & \textbf{60.66}
            &  &  \textbf{71.18} & \textbf{69.78}    

                \\ \bottomrule
\end{tabular}
}
\caption{Performance comparison between our proposed BeMERC and existing MERC methods on MELD dataset. The experimental results of all the comparative experiments are derived from previously published articles.  LLMERC (only use MERC Instruction Tuning stage) which we propose is the baseline for multimodal emotion recognition in conversation based on large
language model. The best results are bolded compared with the baseline LLMERC. `-' means that the results are unavailable from the original paper.}
\label{tab2}

\end{table*}

%% file: table4.tex

\begin{table}[t!]
\renewcommand{\arraystretch}{1}
\setlength{\tabcolsep}{7pt}  
\centering
\resizebox{1.0\linewidth}{!}
{
\begin{tabular}{l|cccccc}
\toprule
\multirow{2}{*}{} &  &                                                                                                                                        
                         \multicolumn{2}{c}{IEMOCAP} &  & \multicolumn{2}{c}{MELD} \\
                          &  & Acc.           & w-F1  
                          &  & Acc.           & w-F1            \\ \midrule

BeMERC                  &  & 74.98 & 74.88       &  & 71.18 & 69.78  \\
\hdashline


w/o B+P                  &  & 74.74 & 74.53       &  & 70.34 & 69.52    \\
w/o P+F                 &  & 74.86 & 74.53       &  & 70.57 & 69.48    \\
w/o B+F                  &  & 73.32 & 73.10       &  & 70.45 & 69.10    \\

w/o  F+B+P                &  & 71.66 & 71.51       &  & 70.22 & 68.90        

                \\ \bottomrule
\end{tabular}
}
\caption{Results of using different behavior. `F', `B' and `P' respectively refer to `facial expression', `body language' and `posture'.}
\label{tab4}
\end{table}

%% file: leida.tex
\begin{figure}[t]
    \centering
  \includegraphics[width=1\linewidth]
  {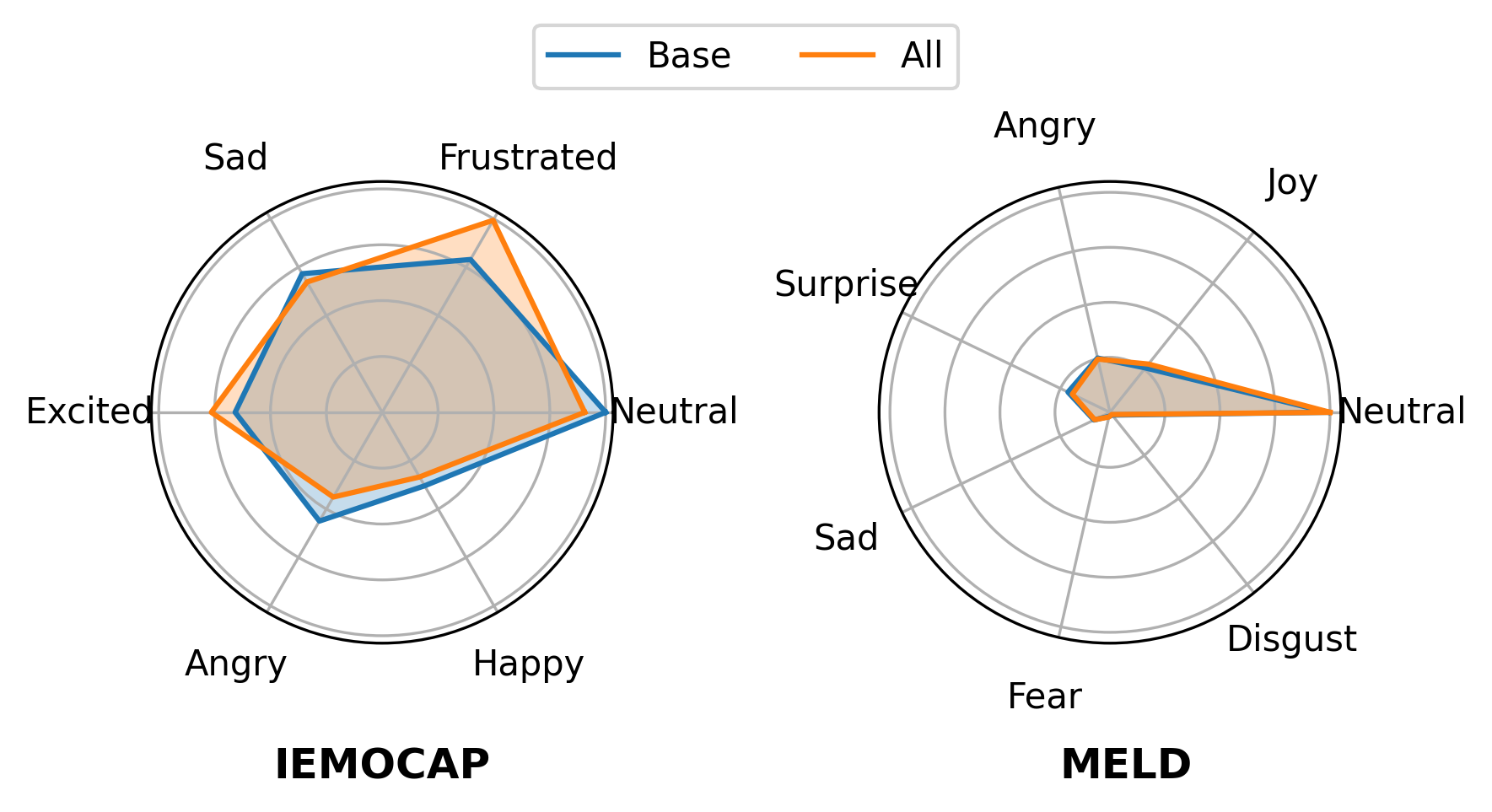}
  \caption {Predicted number of labels for each category on two datasets. `Base' refers to `LLMERC'. `All' refers to `BeMERC'.}
  \label{leida}
\end{figure}

%% file: leida2.tex
\begin{figure}[t]
    \centering
  \includegraphics[width=1\linewidth]
  {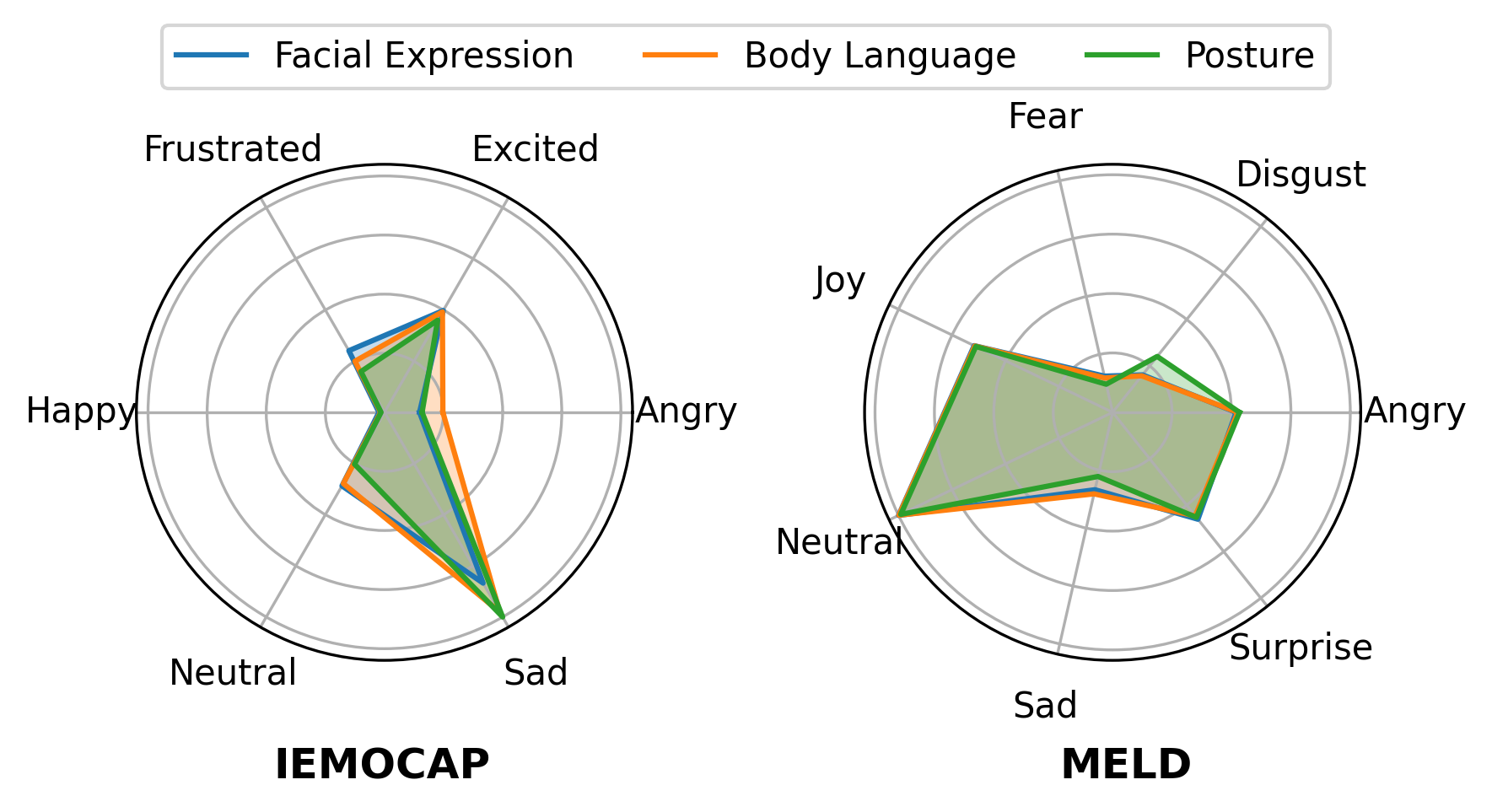}
  \caption {Performance (F1) of each component of 
BeMERC across different emotional category on two datasets.}
  \label{leida2}
\end{figure}

%% file: julei.tex
\begin{figure}[t]
    \centering
  \includegraphics[width=1\linewidth]
  {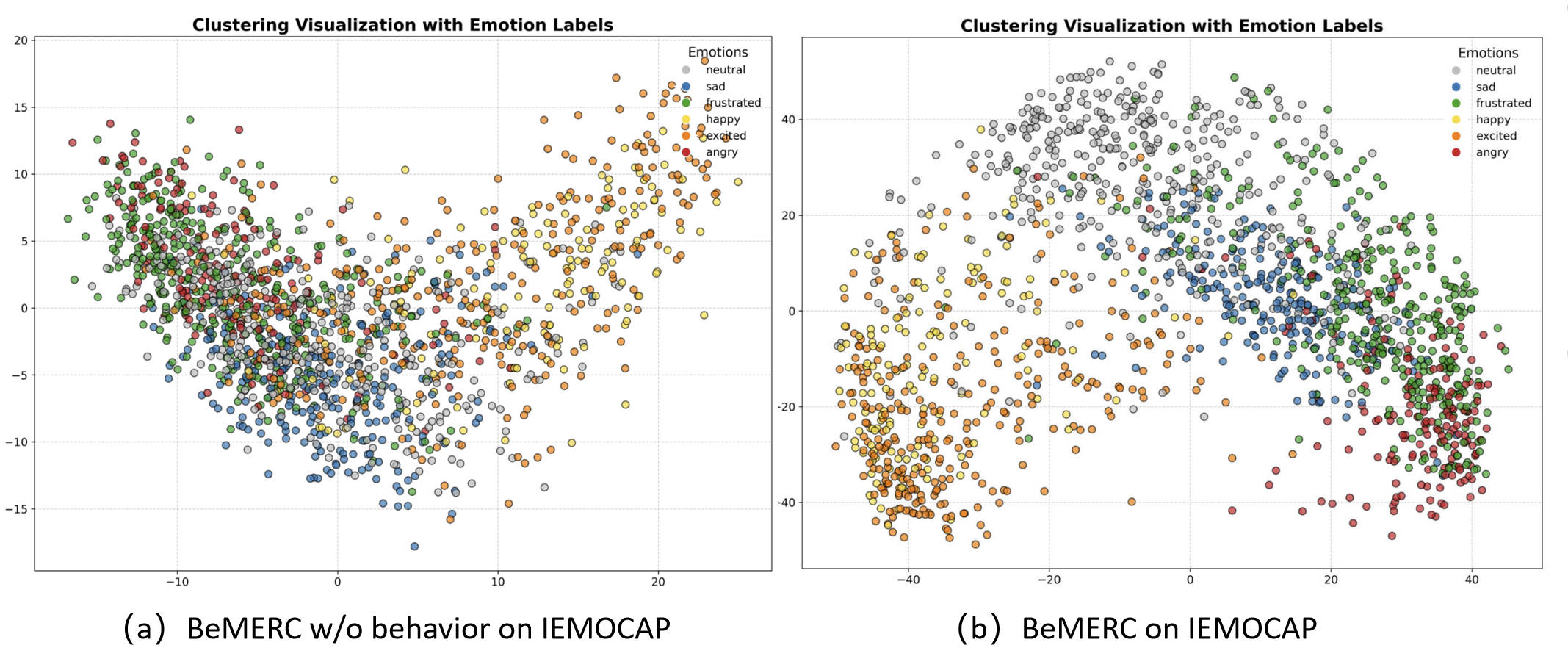}
  \caption {Visualization of thelearned embeddings.}
  \label{julei}
\end{figure}

%% file: table6.tex

\begin{table}[t!]
\renewcommand{\arraystretch}{1}
\setlength{\tabcolsep}{3pt}  
\centering
\resizebox{1.0\linewidth}{!}
{
\begin{tabular}{l|cccccc}
\toprule
\multirow{2}{*}{} &  &                                                                                                                                        
                         \multicolumn{2}{c}{zero-shot} &  & \multicolumn{2}{c}{zero-shot + Behavior} \\
                          &  & Acc.           & w-F1  
                          &  & Acc.           & w-F1            \\ \midrule

GPT-4o-mini                  &  & 53.72 & 51.39       &  & 54.09 & 52.73    \\ 
Gemini                  &  & 55.63 & 55.10       &  & 55.88 & 55.28    \\
Qwen-2.5-72B                 &  & 56.07 & 54.77       &  & 59.03 & 57.92    \\
Phi-4-14B                  &  & 55.21 & 53.20       &  & 55.67 & 55.18    

                \\ \bottomrule
\end{tabular}
}
\caption{Results in a zero-shot scenario on IEMOCAP.}
\label{tab6}
\end{table}